\documentclass[10pt,twocolumn,letterpaper]{article}
\usepackage[pagenumbers]{cvpr} 
\usepackage{graphicx}
\usepackage{amsmath}
\usepackage{amssymb}
\usepackage{booktabs}
\usepackage[pagebackref=true,breaklinks=true,colorlinks,bookmarks=false]{hyperref}
\usepackage{url}            
\usepackage{booktabs}       
\usepackage{amsfonts}       
\usepackage{nicefrac}       
\usepackage{caption}
\usepackage{graphicx}
\usepackage{amssymb}
\usepackage{pifont}
\usepackage{amsmath}
\usepackage{multirow}
\usepackage{bm}
\usepackage[dvipsnames]{xcolor}
\usepackage{times,amsmath,amssymb,booktabs,tabulary,multirow,overpic,xcolor}
\usepackage{epsfig}

\definecolor{citecolor}{RGB}{34,139,34}
\usepackage{colortbl} 

\makeatletter\renewcommand\paragraph{\@startsection{paragraph}{4}{\z@}
      {.5em \@plus1ex \@minus.2ex}{-.5em}{\normalfont\normalsize\bfseries}}\makeatother

\newcommand{\bd}[1]{\textbf{#1}}
\newcommand{\app}{\raise.17ex\hbox{$\scriptstyle\sim$}}

\newcolumntype{x}[1]{>{\centering\arraybackslash}p{#1pt}}

\newlength\savewidth\newcommand\shline{\noalign{\global\savewidth\arrayrulewidth
  \global\arrayrulewidth 1pt}\hline\noalign{\global\arrayrulewidth\savewidth}}
\newcommand{\tablestyle}[2]{\setlength{\tabcolsep}{#1}\renewcommand{\arraystretch}{#2}\centering\footnotesize}

\makeatletter
\newcommand{\printfnsymbol}[1]{%
  \textsuperscript{\@fnsymbol{#1}}%
}

\usepackage{bbm}
\definecolor{mygray}{gray}{0.6}
\definecolor{mygray-bg}{gray}{0.95}
\newcommand{\xmark}{\ding{55}}%
\usepackage[capitalize]{cleveref}
\crefname{section}{Sec.}{Secs.}
\Crefname{section}{Section}{Sections}
\Crefname{table}{Table}{Tables}
\crefname{table}{Tab.}{Tabs.}

\def\cvprPaperID{2258} 
\def\confName{CVPR}
\def\confYear{2022}

\begin{document}

\title{GEN-VLKT: Simplify Association and Enhance Interaction Understanding \\ for HOI Detection}

\author{  
  Yue Liao\textsuperscript{\rm 1}\thanks{Equal contribution} \quad Aixi Zhang\textsuperscript{\rm 2}\printfnsymbol{1} \quad Miao Lu\textsuperscript{\rm 2} \quad Yongliang Wang\textsuperscript{\rm 2} \quad Xiaobo Li\textsuperscript{\rm 2}  \quad Si Liu\textsuperscript{\rm 1}\thanks{Corresponding author (liusi@buaa.edu.cn)} \\

   \textsuperscript{\rm 1}Beihang University \quad
  \textsuperscript{\rm 2}Alibaba Group 
}
\maketitle

\begin{abstract}
The task of Human-Object Interaction~(HOI) detection could be divided into two core problems, i.e., human-object association and interaction understanding. In this paper, we reveal and address the disadvantages of the conventional query-driven HOI detectors from the two aspects.
For the association, previous two-branch methods suffer from complex and costly post-matching, while single-branch methods ignore the features distinction in different tasks. We propose Guided-Embedding Network~(GEN) to attain a two-branch pipeline without post-matching. In GEN, we design an instance decoder to detect humans and objects with two independent query sets and a position Guided Embedding~(p-GE) to mark the human and object in the same position as a pair. Besides, we design an interaction decoder to classify interactions, where the interaction queries are made of instance Guided Embeddings~(i-GE) generated from the outputs of each instance decoder layer.
For the interaction understanding, previous methods suffer from long-tailed distribution and zero-shot discovery. This paper proposes Visual-Linguistic Knowledge Transfer~(VLKT) training strategy to enhance interaction understanding by transferring knowledge from a visual-linguistic pre-trained model CLIP.  In specific, we extract text embeddings for all labels with CLIP to initialize the classifier and adopt a mimic loss to minimize the visual feature distance between GEN and CLIP. As a result, GEN-VLKT outperforms the state of the art by large margins on multiple datasets, e.g., $+5.05$ mAP on HICO-Det. The source codes are available at \url{https://github.com/YueLiao/gen-vlkt}.
\end{abstract}

\begin{figure}[h]
  \centering
  \includegraphics[width=1\linewidth]{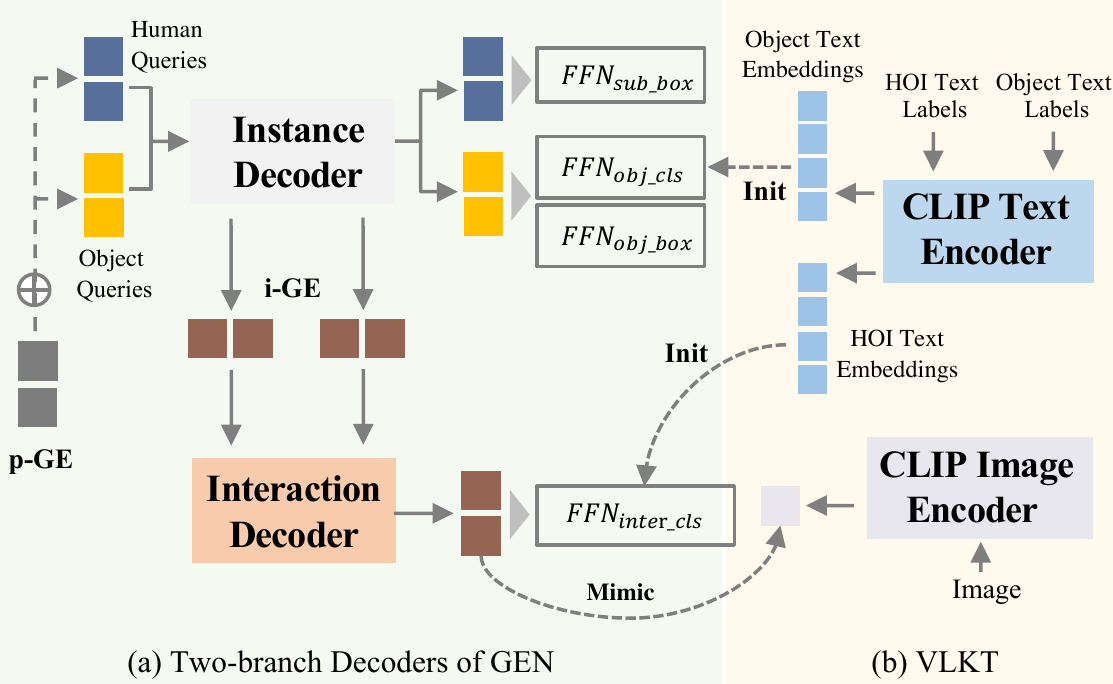}
  \vspace{-4.5mm}
  \caption{\textbf{Our GEN-VLKT pipeline.} We propose GEN, a query-based HOI detector with two-branch decoders, where we design a guided embedding association mechanism to replace the traditional post-matching process for simplifying the association. Moreover, we devise a training strategy VLKT, where we transfer knowledge from the large-scale visual-linguistic pre-trained model CLIP to enhance interaction understanding.}
  \label{fig:first}
   \vspace{-3mm}
\end{figure}

\vspace{-4mm}\section{Introduction}
\vspace{-1.5mm}
\label{sec:intro}

Human-Object Interaction~(HOI) detection is a significant task to make a machine understand human activities in a static image at a fine-grained level. In this task, human activities are represented as a series of HOI triplets~$<$Human, Object, Verb$>$, so an HOI detector is required to localize human and object pairs and recognize their interactions. The core problems of HOI detection are to explore how to \emph{associate the interactive human and object pairs} and \emph{understand their interactions}. Thus, we consider improving the HOI detector from the two aspects and design a unified and superior HOI detection framework. We first revisit the efforts conducted by traditional methods in such two aspects.

For the \emph{association problem}, it can be mainly divided into two paradigms, \emph{i.e.}, bottom-up and top-down. Bottom-up methods~\cite{li2018transferable,gao2018ican,Gao-ECCV-DRG} detect humans and objects first and then associate humans and objects through a classifier or a graph model. Top-down methods usually design an anchor to denote the interaction, \emph{e.g.}, interaction point~\cite{liao2020ppdm} and queries~\cite{tamura2021qpic,zou2021_hoitrans,chen_2021_asnet}, and then find the corresponding human and object through pre-defined associative rules.  Benefiting from the development of visual transformer, query-based methods are leading the performance of HOI detection, which are mainly two streams, \emph{i.e.}, two-branch prediction-then-matching manner~\cite{chen_2021_asnet} and single-branch directly-detection manner~\cite{tamura2021qpic,zou2021_hoitrans}. The two-branch manner predicts interaction then matches with human and object, struggling with designing effective matching rules and complicated post-processing. The single-branch manner proposes to detect the human, object and the corresponding interaction based on a single query with multiple heads in an end-to-end manner. However, we argue that the three tasks, \emph{i.e.}, human detection, object detection and interaction understanding, exist significant differences in feature representation, where human and object detection mainly focus on the features in their corresponding regions, while interaction understanding attends human posture or context. 

To improve this, as shown in Figure~\ref{fig:first}a, we propose to keep the two-branch architecture while removing the complicated post-matching. To this end, we propose Guided Embedding Network~(GEN), where we adopt an architecture of a visual encoder followed by two-branch decoders, \emph{i.e.}, instance decoder and interaction decoder, and design a guided embedding mechanism to guide the association beforehand. The two branches are both with a query-based transformer decoder architecture. For the instance decoder, we design two independent query sets for human and object detection. Further, we develop a position Guided Embedding~(p-GE) to distinguish different human-object pairs by assigning the human query and object query at the same position as a pair. For the interaction decoder, we devise an instance Guided Embedding~(i-GE), where we generate each interaction query guided by specific human and object queries to predict its HOIs. Hence, GEN can allow different features for different tasks and guide the association during network forward while without post-matching.

For the \emph{interaction understanding problem}, most conventional methods directly apply a multi-label classifier fitted from the dataset to recognize the HOIs.
However, such paradigms suffer from the long-tailed distribution and zero-shot discovery due to the complicated human activities with various interactive objects in realistic scenes. Though recent methods propose to alleviate such problems with data-augmentation~\cite{hou2021affordance} or carefully designed loss~\cite{zhong2021glance}, the performance gain and extension ability are restricted to the limited training scale due to the expensive HOI annotation. We might as well set our sights on image-text data, which can be easily obtained from the internet, while we can naturally convert HOI triplets into text descriptions. Thanks to the development of visual-linguistic pre-trained models, this becomes possible. Especially, CLIP~\cite{radford2021learning} establishes a strong visual-linguistic model trained on about 400 million image-text pairs and shows its powerful generalization ability on about $30$ tasks. Thus, CLIP can cover most HOI scenes in real life and bring a new idea to understand the HOIs. 

To improve this, as shown in Figure~\ref{fig:first}b, we design a Visual-Linguistic Knowledge Transfer~(VLKT) training strategy to transfer the knowledge from CLIP to the HOI detector to enhance interaction understanding without additional computation cost. We consider two main problems in our VLKT. On the one hand, we design a text-driven classifier for prior knowledge integration and zero-shot HOI discovery. In detail, we first covert each HOI triplet label into a phrase description, then extract their text embeddings based on the text encoder of CLIP. Finally,  we apply the text embeddings of all HOI labels to initialize the weight of the classifier. In this manner, we can easily extend a novel HOI category only by adding its text embedding into the matrix. Meanwhile, we also adopt the CLIP-initialized object classifier for novel object extension. On the other hand, for text-driven classifier and visual feature alignment, we present a knowledge distillation method to guide the visual features of HOI detection to mimic the CLIP features. Therefore, based on VLKT, the model can well capture information from CLIP and easily extend to novel HOI categories without extra cost during inference.

Finally, we propose a novel unified HOI detection framework GEN-VLKT based on the above two designs. We have verified the effectiveness of our GEN-VLKT on two representative HOI detection benchmarks, \emph{i.e.}, HICO-Det~\cite{qi2018learning}
 and V-COCO~\cite{gupta2015visual}. Our GEN-VLKT has significantly improved the existing methods on both two benchmarks and achieved state-of-the-arts on the zero-shot settings of the HICO-Det dataset. Specifically, our GEN-VLKT has achieved a $5.05$ mAP gain on HICO-Det and a $5.28$ AP promotion on V-COCO compared with the previous state-of-the-art method QPIC~\cite{tamura2021qpic}. It also promotes performance impressively by a $108.12\%$ relative mAP gain for unseen object zero-shot setting compared to the previous state-of-the-art method ATL~\cite{hou2021affordance}.

\begin{figure*}[t]
\begin{center}
   \includegraphics[width=1.0\linewidth]{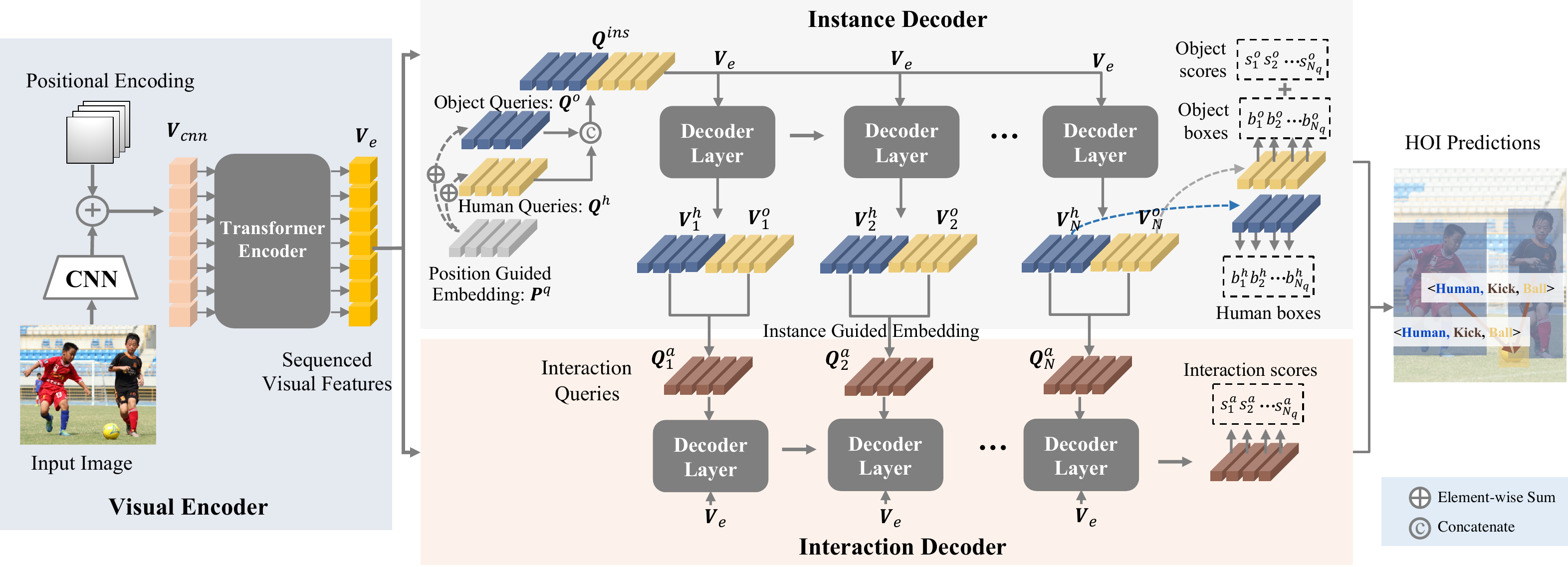}
\end{center}
\vspace{-4mm}
   \caption{\bd{The framework of our GEN}. The GEN is organized as a visual encoder equipped with two-branch decoders architecture. Given an image, the visual encoder is first applied to extract the visual features. Then, two branches, \emph{i.e.}, instance decoder and interaction decoder, are used to localize human-object pairs and classify HOI triplets based on learnable queries, respectively. Besides, we design a position Guided Embedding~(p-GE) to associate the interactive human and object and an instance Guided Embedding~(i-GE) to make the interaction query predict the corresponding HOI categories under the guidance of specific human and object queries.}
   \label{fig:gen}
  \vspace{-3mm}
\label{fig:2}
\end{figure*}
 
\vspace{-2mm}\section{Related Works}
\vspace{-1mm}\noindent\textbf{HOI detection.} Conventional HOI detectors are mainly divided into two folds, bottom-up and top-down.
The bottom-up pipelines~\cite{gao2018ican,chao2018learning,li2018transferable,Gupta_2019_ICCV,Zhou_2019_ICCV,Wan_2019_ICCV,li2020detailed,Xu_2019_CVPR,Gao-ECCV-DRG,Liu20a,kim2020detecting,li2021improving} first detect all humans and objects and then associate the human-object pairs and infer their HOI types through an additional classifier. These methods are usually organized as a two-stage paradigm and worked on improving the second stage. Recently, some graph-based methods~\cite{qi2018learning,Ulutan_2020_CVPR,wang2020contextual,YangZ20_IJCAI_In-GraphNet,Zhou_2019_ICCV} have achieved satisfactory performance. However, bottom-up methods suffer from expensive computation consume due to its serial architecture for handling a large number of human-object pairs. 
To alleviate this issue, top-down methods become popular in recent works~\cite{liao2020ppdm,wang2020learning,Kim2020_unidet,zou2021_hoitrans,chen_2021_asnet,tamura2021qpic,Kim_2021_CVPR}. 
Top-down methods mainly design an additional anchor to associate humans and objects, and predict their interactions. The interaction anchor is from the early interaction point~\cite{liao2020ppdm,wang2020learning} and union box~\cite{Kim2020_unidet} to recent interaction query~\cite{chen_2021_asnet,tamura2021qpic,zou2021_hoitrans,Kim_2021_CVPR} with the development of visual transformers. Recently, CDN~\cite{zhang2021mining} proposed a one-stage method with a cascade decoder to mine the benefits of the two-stage and one-stage HOI detectors. Our GEN is different from CDN in the three aspects. 1) Organization of the decoder: GEN is with a two-branch pipeline and the instance and interaction decoder forward together,
while CDN disentangles the HOI detection into two decoders in a serial manner. 2) Instance query design: GEN adopts two isolated human and object queries with positional embedding, while CDN entangles human and object into a unified instance query. 3) motivation: GEN aims to replace the complex post-process with a guided learning manner, while CDN aims to mine the benefits of one- and two-stage detectors.

\noindent\textbf{Zero-shot HOI detection.} Zero-shot HOI detection~\cite{shen2018scaling} tends to detect unseen HOI triplet categories in the training data. Many methods~\cite{shen2018scaling,gupta2019no,ulutan2020vsgnet,Bansal2020_aaai_functional,hou2020visual,hou2021detecting,hou2021affordance,xu2019learning,peyre2019detecting,liu2020consnet} are investigated to handle zero-shot HOI detection. In detail, ~\cite{shen2018scaling,gupta2019no,ulutan2020vsgnet,Bansal2020_aaai_functional} factorized the human and object features by disentangled reasoning on verbs and objects and then produced novel HOI triplets during inference. VCL~\cite{hou2020visual} composed novel HOI samples by combining decomposed object and verb features with pair-wise images and within images. FCL~\cite{hou2021detecting} presented an object fabricator to generate fake object representations for rare and unseen HOIs. ATL~\cite{hou2021affordance} explored object affordances from additional object images to discover novel HOI categories. ConsNet~\cite{liu2020consnet} explicitly encoded the relations among objects, actions and interactions into an undirected graph to propagate knowledge among HOI categories as well as their constituents. The visual-linguistic models~\cite{xu2019learning,peyre2019detecting} transferred the seen visual phrase embeddings with prior language knowledge to unseen HOIs. 

\noindent\textbf{HOI Detection with Visual-linguistic Model.} An HOI triplet can be regarded as a structural linguistic description, so integrating linguistic prior knowledge into HOI detection is a natural idea. Most conventional methods~\cite{Gao-ECCV-DRG,Kim_2021_CVPR,zhong2020polysemy,peyre2019detecting} focus on mining the inductive bias as a pair-wise frequency prior manner or just utilizing the word embeddings as additional features. Recently, large-scale pre-trained visual-linguistic models~\cite{radford2021learning,su2019vl,lu2019vilbert} have shown their powerful performances and generalization abilities in many visual or cross modality tasks. Recently, DEFR~\cite{jin2021object} proposed to adopt a visual-linguistic model, CLIP, to initialize the classier for the HOI recognition task.  We aim to explore applying visual-linguistic pre-trained model for interaction understanding.

\vspace{-2mm}
\section{Methods}
\vspace{-1mm}
In this section, we aim to explore the solutions for the two problems of HOI detection, \emph{i.e.}, association and interaction understanding. We first present a detailed introduction of our one-stage two-branch HOI detector with a simple association mechanism, namely Guided Embedding Network~(GEN), in Sec~\ref{sec:pen}. We then introduce a Visual-Linguistic Knowledge Transfer~(VLKT) training strategy with the large-scale visual-linguistic pre-trained model CLIP to enhance interaction understanding in Sec~\ref{sec:clip}. 
Finally, we show the training and inference pipelines.

\vspace{-1mm}\subsection{Guided Embedding Network}\label{sec:pen}\vspace{-1mm}
In this subsection, we introduce the architecture of our Guided Embedding Network~(GEN). As shown in Figure~\ref{fig:gen}, the GEN is organized as an encoder followed by two-branch decoders architecture. We first adopt a CNN equipped with a transformer encoder architecture as the visual encoder to extract sequenced visual features $\bm{V}_e$. Then, we apply two-branch decoders, \emph{i.e.}, instance decoder and interaction decoder, to detect HOI triplets. In the instance decoder, based on $\bm{V}_e$, we detect humans and objects through the human query set $\bm{Q}^h$ and the object query set $\bm{Q}^o$ individually. Additionally, we design a position Guided Embedding~(p-GE) $\bm{P}^q$ to assign the human and object queries at the same position as a pair. In the interaction decoder, we first dynamically generate the interaction queries $\bm{Q}^a_i$ for each interaction decoder layer by computing the mean of the outputs of human and object queries in the corresponding instance decoder layer. Therefore, the interaction decoder can predict the corresponding HOI categories under the guidance of human and object queries. Finally, the HOI prediction results are generated by the output of decoders.

\vspace{1mm}\noindent\textbf{Visual Encoder.} We follow the query-based transformer detectors~\cite{carion2020endtoend,tamura2021qpic,zou2021_hoitrans} to adopt a CNN-transformer combined architecture for the visual encoder. Taking an image $\bm{I}$ as input, a CNN is first utilized to extract low-resolution visual features $\bm{V}_{cnn} \in \mathbb{R}^{H'\times W' \times C'}$. Then, we reduce the channels of visual features to $C_e$ and flatten the size of the features to ${(H'\times W') \times C_e}$. Finally, we feed the reduced features adding a cosine positional embedding into a transformer encoder, and extract sequenced visual features $\bm{V}_e \in \mathbb{R}^{(H'\times W') \times C_e}$ for the following tasks.

\vspace{1mm}\noindent\textbf{Two-branch Decoders.} The decoders in the two branches share the same architecture, where we follow the transformer-based detectors~\cite{carion2020endtoend,chen_2021_asnet} to adopt the query-based transformer decoder framework. First, we feed a set of learnable queries $\bm{Q}\in \mathbb{R}^{N_q\times C_q}$, the output of last layer, the visual features $V_e$ and the positional embedding to $N$ transformer decoder layers, and output the updated queries after the self-attention and co-attention operations. Then with separate FFN heads, the queries are transformed to embeddings for its dedicated task, \emph{i.e.}, instance and interaction representations by the first and second decoder branch, respectively.

\begin{figure}[t]
\begin{center}
   \includegraphics[width=1.0\linewidth]{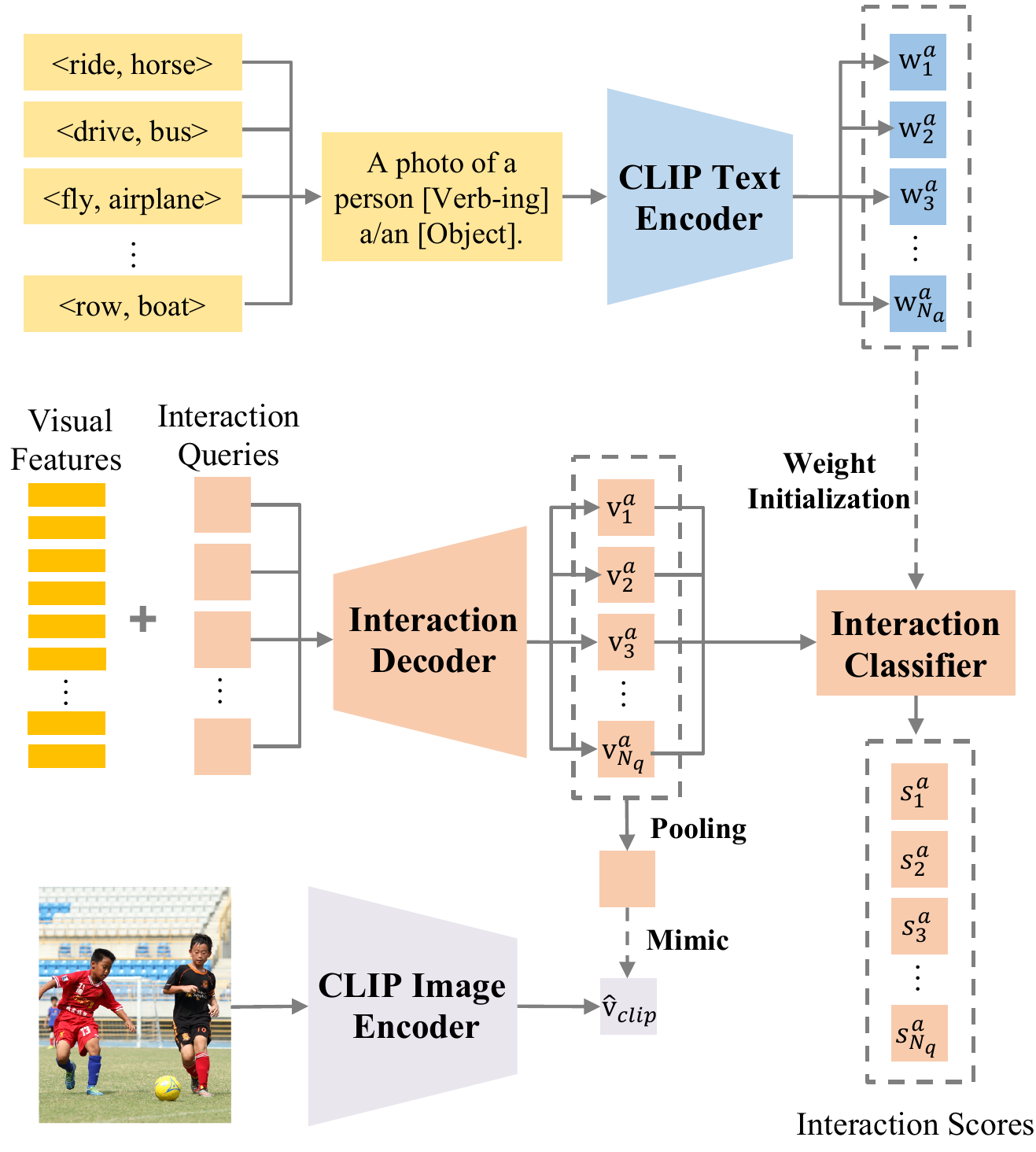}
\end{center}
\vspace{-3.5mm}
   \caption{\bd{VLKT for the interaction decoder}. We first covert each HOI label into a phrase description and extract its text embedding based on CLIP text encoder, then apply the text embeddings of all HOI labels to initialize the classifier. Finally, we adopt the CLIP image encoder to extract visual features to guide the interaction visual feature learning. The dotted arrow denotes no gradient.}
  \vspace{-3.5mm}
\label{fig:clip}
\end{figure}

For the instance decoder, we first initialize two sets of queries to detect human and object ~\cite{dong2021visual} where we denote human and object query sets as $\bm{Q}^h\in \mathbb{R}^{N_q\times C_q}$ and $\bm{Q}^o\in \mathbb{R}^{N_q\times C_q}$ separately. Then, we design an additional learnable position Guided Embedding~(p-GE) $\bm{P}^q \in \mathbb{R}^{N_q\times C_q}$ for the two query sets to assign the human query and object query at the same position as a pair, where we add the $\bm{P}^q$ to $\bm{Q}^h$ and $\bm{Q}^o$, respectively. Finally, we generate the query set for the instance decoder by concatenating the added queries:
\begin{equation}
\small
\begin{aligned}
\bm{Q}^{ins} = [\bm{Q}^h + \bm{P}^q, \bm{Q}^o + \bm{P}^q],
\end{aligned}
\end{equation}
where $\bm{Q}^{ins} \in \mathbb{R}^{2N_q\times C_q}$. We feed $\bm{Q}^{ins}$ forward the instance decoder to predict the human-object bounding-box pairs $(\bm{b}^h_i, \bm{b}^o_i, \bm{s}^o_i)$, where $\bm{b}^h_i \in \bm{B}^h$,  $\bm{b}^o_i \in \bm{B}^o$ and $\bm{s}^o_i \in \bm{S}^o$ denote human bounding-box, object bounding-box and object category scores. And we extract the middle features decoded by each decoder layer as $\mathbf{V}^{ins} = [\mathbf{V}^{h}, \mathbf{V}^{o}]$ for the following interaction decoder, where $\mathbf{V}^{ins} \in \mathbb{R}^{N \times 2N_q\times C_q}$.

The goal of the interaction decoder is to predict the HOI categories for the corresponding human-object pair. Therefore, this branch is required to associate interaction query with the human-object query pairs and classify interaction. Here, we introduce an instance Guided Embedding~(i-GE) method for the association, and the new interaction classification manner will be introduced in the next subsection.
Instead of conventional learnable embedding with random initialization, we dynamically generate i-GE as the interaction queries to guide the interaction query to match the human and object queries.
In this manner, we generate i-GE under the guidance of the middle visual features $[\mathbf{V}^{h}, \mathbf{V}^{o}]$. Specifically, for the input of $k$-th layer interaction decoder, the interaction queries $\bm{Q}^{a}_k$ is computed by the outputs of $k$-th layer instances decoder:
\begin{equation}
\small
\begin{aligned}
\bm{Q}^{a}_k = (\bm{V}^h_k + \bm{V}^o_k) / 2.
\end{aligned}
\end{equation}
In this way, the $k$-th layer interaction decoder takes the interaction queries  $\bm{Q}^{a}_k$ as input, and return the immediate decoded features $\bm{V}^a_k$ and HOI categories.

\vspace{-1mm}\subsection{Visual-Linguistic Knowledge Transfer}\label{sec:clip}\vspace{-1mm}
In this subsection, we detailedly introduce the training pipeline of the instance decoder and the interaction decoder transferring knowledge from the large-scale visual-linguistic pre-trained model CLIP~\cite{radford2021learning}, namely Visual-Linguistic Knowledge Transfer~(VLKT). In VLKT, inspired by~\cite{jin2021object,gu2021open}, we first adopt the CLIP text embeddings to classify interactions and objects.
 We then introduce how to transfer the visual knowledge from CLIP image embedding to the interaction decoder. We present the pipeline of interaction decoder training with VLKT in Figure~\ref{fig:clip}.

\vspace{1mm}\noindent\textbf{Text Embedding for Classifier Initialization.} To generate the CLIP text embedding, we first convert HOI triplet labels and object labels into text descriptions. For example, given an HOI triplet $<$Human, Object, Verb$>$, we generate the corresponding description following such format `A photo of a person [Verb-ing] a/an [Object]'. In addition, the `no-interaction' type is represented as `A photo of a person and a/an [Object]'. As for an object label, we transform it into the phrase `A photo of a/an [Object]'. Then, we generate the text embedding for each HOI and object text label through the pre-trained CLIP text-encoder offline. Finally, the text embedding set of HOI labels $\bm{E}^a \in \mathbb{R}^{c_{t} \times N_a}$ and object labels $\bm{E}^o \in \mathbb{R}^{c_{t} \times N_o}$ are obtained, where $N_a$ and $N_o$ denote the number of HOI triplet categories and object categories, respectively, and $c_t$ represents the dimension of text embedding from CLIP text encoder.

After obtaining the text embeddings, we aim to classify the interaction and object under the guidance of the prior knowledge from such text embeddings. The intuitive idea is to adopt such embeddings to initialize the weight of the learnable classifier and fine-tune the classifier with a small learning rate to fit a specific dataset. In this way,  each output query feature is computed cosine similarity with all fine-tuned text embeddings and returned a similarity score vector during the classification process. Specifically, we denote the interaction classifier and object classifier as $\mathcal{C}^a \leftarrow \bm{E}^a$ and $\mathcal{C}^o \leftarrow \bm{E}^o$, respectively. Taking interaction classifier $\mathcal{C}^a = [\mathbf{w}^a_1, \mathbf{w}^a_2,...,\mathbf{w}^a_{N_a}]$ as an example, given an output interaction query $\mathbf{v}_i^a$, we compute the similarity score by:
\begin{equation}
\small
\begin{aligned}
\mathbf{s}^a_i =\theta\left[\operatorname{sim}\left(\mathbf{v}_{i}^a, \mathbf{w}^a_1\right), \operatorname{sim}\left(\mathbf{v}_{i}^a, \mathbf{w}^a_2\right), \cdots, \operatorname{sim}\left(\mathbf{v}_{i}^a, \mathbf{w}^a_{N_a}\right)\right] 
\end{aligned}
\end{equation}
where $\operatorname{sim}$ denotes the cosine similarity operation, for example $\operatorname{sim}\left(\mathbf{v}_{i}^a, \mathbf{w}^a_1\right) = (\mathbf{v}_{i}^a \cdot \mathbf{w}^a_1) /( \|\mathbf{v}_{i}^a \|\|\mathbf{w}^a_1\|)$, and $\theta$ is a logit scale factor following CLIP~\cite{radford2021learning}. The object classification scores can be got in the same way. Otherwise, we follow~\cite{tamura2021qpic} to apply the focal loss and cross-entropy loss to train the interaction and object classifier, respectively.

\vspace{1mm}\noindent\textbf{Visual Embedding Mimic.}
 CLIP is trained on image-text pair data, and it aligns visual embedding and text embedding into a unified space. We design a Visual Embedding Mimicking mechanism to pull the interaction feature into such unified space by pulling the distance between the interaction feature and CLIP visual embedding. 
 Here, CLIP serves as the teacher, and the interaction decoder plays the student's role. We design the knowledge distillation strategy from the global image level, because CLIP image encoder is built upon a whole image. We first feed the resized and cropped image into the pre-trained CLIP visual encoder and extract the visual embedding $\hat{\mathbf{v}}_{clip}$ for the teacher supervision. The global student visual embedding is generated by conducting an average pooling among all output interaction query features. $\mathcal{L}_1$ loss is utilized to pull the distance between the student and the teacher. We formulate the global knowledge distillation as:
\begin{equation}
\small
\begin{aligned}
\mathcal{L}_{glo} =|\hat{\mathbf{v}}_{clip}  - \frac{1}{N_q}\sum_{i=1}^{N_q}\mathbf{v}_{i}^a|, 
\end{aligned}
\end{equation}
where $N_q$ denotes the number of queries.

\vspace{-1mm}\subsection{Training and Inference} \label{sec:train}\vspace{-1mm}
In this subsection, we elaborate the processes of training and inference.

\vspace{1mm}\noindent\textbf{Training.} During the training stage, we follow the query-based methods~\cite{carion2020endtoend,zou2021_hoitrans,tamura2021qpic} to assign a bipartite matching prediction with each ground-truth using the Hungarian algorithm. The matching process combines the predictions from the FFN heads of the two-branch decoders since the queries of human, object and interaction are one-to-one corresponding. The matching cost for the matching process and the targeting cost for the training back-propagation share the same strategy, where is composed by the box regression loss $\mathcal{L}_b$, the intersection-over-union loss $\mathcal{L}_u$ and the classification loss $\mathcal{L}_c$. The cost is formulated as:
\begin{equation}
\mathcal{L}_{cost} = \lambda_b \sum_{i\in(h,o)} \mathcal{L}_b^i + \lambda_u \sum_{j\in(h,o)} \mathcal{L}_u^j + \sum_{k\in(o,a)} \lambda_c^k \mathcal{L}_c^k,
\label{eq02}
\end{equation}
where $\lambda_b$, $\lambda_u$ and $\lambda_c^k$ are the hyper-parameters for adjusting the weights of each loss. Then, considering the mimic loss, the final training loss is given as:
\begin{equation}
\mathcal{L} = \mathcal{L}_{cost} + \lambda_{mimic} \mathcal{L}_{glo},
\label{eq03}
\end{equation}
where $\lambda_{mimic}$ is the hyper-parameter weight for distilling the image embeddings. Additionally, we apply an intermediate supervision for the output of each decoder layer.

\vspace{1mm}\noindent\textbf{Inference.} The visual embedding mimic only contributes to the training stage, removing it during inference. For each human-object bounding-box pair ($b_i^h$, $b_i^o$) with the object score $s_i^o$ from instance decoder branch, the interaction score is predicted as $s_i^a$ from the interaction decoder. Then, we extend $s_i^o$ from $N_o$-dim to $N_a$-dim, where the score for a specific object category will be copy-paste several times for all the corresponding HOI categories. The HOI triplet score is given as $ s_i^a + s_i^o(N_a) s_i^o(N_a) $ instead of $ s_i^a s_i^o(N_o) $ for balancing the weights of interaction score and object score. The HOI triplets with top $K$ confidence scores are processed by a triplet NMS as the final predictions.

\vspace{-2mm}\section{Experiments}\vspace{-1mm}
In this section, we demonstrate the effectiveness of our designed GEN-VLKT with comprehensive experiments. In Sec~\ref{sec:setting}, we first introduce our experimental settings. Then in Sec~\ref{sec:regular}, we compare our GEN-VLKT with the previous state-of-the-art approaches. Next, we present the superiority of our GEN-VLKT on zero-shot HOI detection in Sec~\ref{sec:zero-shot}. Finally, we conduct the ablation studies in Sec~\ref{sec:ablation}.

\begin{table*}[!t]
  \begin{center}
  \small
  \resizebox{1.0\textwidth}{!}{%
  \begin{tabular}{cccc|ccc|ccc}
    \hline
    &&&&\multicolumn{3}{c|}{Default} & \multicolumn{3}{c}{Know Object} \\
  Method   &Detector &Backbone    &Anchor   & Full & Rare & Non-Rare & Full  & Rare & Non-Rare\\
  \hline
  Bottom-up Methods:        &   &    &       & & & &  & &\\
  InteractNet~\cite{gkioxari2018detecting} &COCO	&ResNet-50-FPN	&\xmark		&9.94	&7.16	&10.77       &-	&-	&-\\
  GPNN~\cite{qi2018learning} &COCO	&Res-DCN-152		&\xmark			&13.11	&9.34	&14.23       &-	&-	&-\\
  iCAN~\cite{gao2018ican}	  &COCO      &ResNet-50		&\xmark				&14.84	&10.45	&16.15       &16.26	&11.33	&17.73\\
  No-Frills~\cite{Gupta_2019_ICCV}	&COCO	&ResNet-152		&\xmark		&17.18	&12.17	&18.68       &-	&-	&-\\ 
  PMFNet~\cite{Wan_2019_ICCV}	&COCO	&ResNet-50-FPN	&\xmark			&17.46	&15.65	&18.00       &20.34	&17.47	&21.20\\ 
  DRG~\cite{Gao-ECCV-DRG}		&COCO	&ResNet-50-FPN	&\xmark				&19.26	&17.74	&19.71		 &23.40	&21.75	&23.89\\
  VCL~\cite{hou2020visual}		&COCO	&ResNet-50	&\xmark &19.43&	16.55&	20.29&	22.00&	19.09&	22.87\\
  VSGNet~\cite{Ulutan_2020_CVPR}	&COCO	&ResNet-152		&\xmark					&19.80	&16.05	&20.91       &-	&-	&-\\ 
  FCMNet~\cite{Liu20a}	&COCO	&ResNet-50		&\xmark				&20.41	&17.34	&21.56       &22.04	&18.97	&23.12\\
  ACP~\cite{kim2020detecting}	&COCO	&ResNet-152		&\xmark				&20.59	&15.92	&21.98       &-	&-	&-\\

  PD-Net~\cite{zhong2020polysemy}	&COCO	&ResNet-152-FPN	&\xmark			&20.81	&15.90	&22.28       &24.78	&18.88	&26.54\\ 
  SG2HOI~\cite{he2021exploiting}	&COCO	&ResNet-50		&\xmark				&20.93& 18.24 &21.78 &24.83 &20.52& 25.32\\
  DJ-RN~\cite{li2020detailed}		&COCO	&ResNet-50		&\xmark			&21.34	&18.53	&22.18       &23.69	&20.64	&24.60\\
  SCG~\cite{zhang2021spatially}		&COCO	&ResNet-50-FPN		&\xmark			&21.85 &18.11 &22.97      &-& -& - \\
  IDN~\cite{li2020hoi} &COCO &ResNet-50	&\xmark  & 23.36 &22.47 &23.63 &26.43 &25.01 &26.85\\
    ATL~\cite{hou2021affordance}	&HICO-Det	&ResNet-50	&\xmark			&23.81& 17.43 &25.72& 27.38 &22.09 &28.96\\
  \hline 
  Top-down Methods:      &  &               &       & & & &  & &\\
  UnionDet~\cite{Kim2020_unidet}	&COCO		&ResNet-50-FPN	&\emph{B}			&17.58	&11.72	&19.33       &19.76	&14.68	&21.27\\
  IP-Net~\cite{wang2020learning}	&COCO	&Hourglass-104	&\emph{P}				&19.56	&12.79	&21.58      	 &22.05	&15.77	&23.92\\
  PPDM-Hourglass~\cite{liao2020ppdm}	& HICO-Det	&Hourglass-104	&\emph{P}			&21.94	&13.97	&24.32       &24.81	&17.09	&27.12\\
  HOI-Trans~\cite{zou2021_hoitrans}	& HICO-Det	&ResNet-50	&\emph{Q}			&23.46& 16.91& 25.41       &26.15& 19.24& 28.22\\ 
  GG-Net~\cite{zhong2021glance}	&HICO-Det	&Hourglass-104	&\emph{P}		&23.47	&16.48&	25.60&	27.36&	20.23&	29.48\\PST~\cite{dong2021visual}	& -	&ResNet-50	&\emph{Q}			&23.93& 14.98& 26.60& 26.42 &17.61& 29.05\\ 
  HOTR~\cite{Kim_2021_CVPR}	& HICO-Det	&ResNet-50	&\emph{Q}			&25.10& 17.34 &27.42& - & - & -\\ 
  AS-Net~\cite{chen_2021_asnet}		& HICO-Det			&ResNet-50	&\emph{Q}			&28.87	&24.25	&30.25 &31.74	&27.07	&33.14  \\
  QPIC-R50~\cite{tamura2021qpic}		& HICO-Det		&ResNet-50	&\emph{Q}			&29.07	&21.85	&31.23 &31.68	&24.14	&33.93  \\
  QPIC-R101~\cite{tamura2021qpic}		& HICO-Det		&ResNet-101	&\emph{Q}			&29.90	&23.92	&31.69 &32.38	&26.06	&34.27  \\
  \cellcolor{mygray-bg}GEN-VLKT$_s$		& \cellcolor{mygray-bg}HICO-Det		&\cellcolor{mygray-bg}ResNet-50	&\cellcolor{mygray-bg}\emph{Q}				&\cellcolor{mygray-bg}33.75	&\cellcolor{mygray-bg}29.25	&\cellcolor{mygray-bg}35.10      &\cellcolor{mygray-bg}36.78	&\cellcolor{mygray-bg}32.75	&\cellcolor{mygray-bg}37.99 \\
  \cellcolor{mygray-bg}GEN-VLKT$_m$		& \cellcolor{mygray-bg}HICO-Det		&\cellcolor{mygray-bg}ResNet-101	&\cellcolor{mygray-bg}\emph{Q}				&\cellcolor{mygray-bg}34.78	&\cellcolor{mygray-bg}\textbf{31.50}	&\cellcolor{mygray-bg}35.77      &\cellcolor{mygray-bg}38.07	&\cellcolor{mygray-bg}\textbf{34.94}	&\cellcolor{mygray-bg}39.01\\
    \cellcolor{mygray-bg}GEN-VLKT$_l$		&	\cellcolor{mygray-bg}HICO-Det	&\cellcolor{mygray-bg}ResNet-101	&\cellcolor{mygray-bg}\emph{Q}				&\cellcolor{mygray-bg}\textbf{34.95}	&\cellcolor{mygray-bg}31.18	&\cellcolor{mygray-bg}\textbf{36.08}       &\cellcolor{mygray-bg}\textbf{38.22}	&\cellcolor{mygray-bg}34.36	&\cellcolor{mygray-bg}\textbf{39.37}\\
  \hline           
  \end{tabular}}
  \end{center}
  \vspace{-3mm}
  \caption{\textbf{Performance comparison on the HICO-Det test set.} We present an additional tag `Anchor' to disgust the interaction anchor types for top-down methods, where the `\emph{B}', `\emph{P}' and `\emph{Q}' denote bounding-box, point and query, respectively.}
  \label{tb:hico}
  \vspace{-3mm}
  \end{table*}
  
\vspace{-1mm}\subsection{Experimental Setting}\label{sec:setting}\vspace{-1mm}
\noindent\textbf{Datasets.} We evaluate our model on two public benchmarks, HICO-Det~\cite{chao2018learning} and V-COCO~\cite{gupta2015visual}. HICO-Det has $47,776$ images ($38,118$ for training and $9,658$ for testing). It contains $600$ classes of HOI triplets constructed by $80$ object categories and $117$ action categories. V-COCO is a subset of COCO dataset and has $10,396$ images ($5,400$ for training and $4,964$ for testing). It has $29$ action categories which includes $4$ body motions without interaction to any objects. It has the same $80$ object categories. Its actions and objects form $263$ classes of HOI triplets.

\vspace{1mm}\noindent\textbf{Data Structure for Zero-Shot.} For zero-shot HOI detection, we conduct experiments on HICO-Det following the setting in~\cite{Bansal2020_aaai_functional}: 1) Unseen Composition~(UC) and 2) Unseen Object~(UO). Specifically, the UC setting indicates the training data contains all categories of object and verb but misses some HOI triplet categories, while the UO setting means the objects in the unseen triplets also do not appear in the training data. We evaluate the $120$ unseen, $480$ seen, and $600$ full categories for the UC setting. Similar to ~\cite{hou2020visual}, the Rare First UC~(RF-UC) selects unseen categories from tail HOIs preferentially, while the Non-rare First UC~(NF-UC) prefers the head categories. For the UO setting, we use the unseen HOIs with $12$ objects unseen among the total $80$ objects and form $100$ unseen and $500$ seen HOIs. Besides, for a more comprehensive demonstration of our method to investigate the novel HOIs, we propose an Unseen Verb~(UV) setting, where we randomly select $20$ verbs from all total $117$ verbs to form $84$ unseen and $516$ seen HOIs. 

\vspace{1mm}\noindent\textbf{Evaluation Metric.} We follow the settings in ~\cite{chao2018learning} to use the mean Average Precision~(mAP) for evaluation. We define a HOI triplet prediction as a true positive if 1) both predicted human and object bounding-boxes have IoU larger than $0.5$ \emph{w.r.t.} the GT boxes; and 2) both predicted HOI categories are accurate. For HICO-Det, we evaluate the three different category sets: all 600 HOI categories (Full), 138 HOI categories with less than $10$ training instances (Rare) and the other 462 HOI categories (Non-Rare). For V-COCO, we report the role mAPs for two scenarios: S$1$ for the $29$ action categories including the $4$ body motions and S$2$ for the $25$ action categories without the no-object HOI categories.

\begin{table}[t]
\small
  \begin{center}
  \begin{tabular}{ccccc}
  \hline
  Method        &Anchor  & $\operatorname{AP}^{S1}_{role}$ & $\operatorname{AP}^{S2}_{role}$ \\
  \hline\hline
  Bottom-up Methods:                  &                       & \\
  InteractNet~\cite{gkioxari2018detecting}		&\xmark				    &40.0  &-\\
  GPNN~\cite{qi2018learning}    &\xmark				    &44.0  &-\\
  iCAN~\cite{gao2018ican}	      		&\xmark				    &45.3  &52.4\\
  TIN~\cite{li2018transferable}      		&\xmark				    &47.8  &54.2\\
  VCL~\cite{hou2020visual}      		&\xmark				    &48.3  &-\\
  DRG~\cite{Gao-ECCV-DRG}				&\xmark		        &51.0  &-\\
  IP-Net~\cite{wang2020learning}		&\xmark				    &51.0  &-\\
  VSGNet~\cite{Ulutan_2020_CVPR}			&\xmark				    &51.8  &57.0\\ 
  PMFNet~\cite{Wan_2019_ICCV}		&\xmark               &52.0  &-\\ 
  PD-Net~\cite{zhong2020polysemy}			&\xmark               &52.6  &-\\
  FCMNet~\cite{Liu20a}				&\xmark		            &53.1  &-\\
  ACP~\cite{kim2020detecting}		&\xmark	&53.23  &-\\
  IDN~\cite{li2020hoi}			&\xmark		            &53.3  &60.3\\
  
  \hline 
  Top-down Methods:     &                                    &  &\\
  UnionDet~\cite{Kim2020_unidet}		&\emph{B}			        &47.5  &56.2\\
  HOI-Trans~\cite{zou2021_hoitrans}		&\emph{Q}			        &52.9  &-\\
  AS-Net~\cite{chen_2021_asnet}		&\emph{Q}		        &53.9  &-\\
  GG-Net~\cite{zhong2021glance} &\emph{P}		        &54.7  &-\\
  HOTR~\cite{Kim_2021_CVPR} &\emph{Q}			        &55.2  & 64.4\\
  QPIC-R50~\cite{tamura2021qpic}		&\emph{Q}			        &58.8  &61.0 \\
  QPIC-R101~\cite{tamura2021qpic}		&\emph{Q}			        &58.3  &60.7 \\
  \cellcolor{mygray-bg}GEN-VLKT$_s$			 	&\cellcolor{mygray-bg}\emph{Q}			        &\cellcolor{mygray-bg}62.41 &\cellcolor{mygray-bg}64.46\\
    \cellcolor{mygray-bg}GEN-VLKT$_m$			 	&\cellcolor{mygray-bg}\emph{Q}			        &\cellcolor{mygray-bg}63.28  &\cellcolor{mygray-bg}65.58\\
    \cellcolor{mygray-bg}GEN-VLKT$_l$			 	&\cellcolor{mygray-bg}\emph{Q}			        &\cellcolor{mygray-bg}\textbf{63.58}  &\cellcolor{mygray-bg}\textbf{65.93}\\
  \hline
  \end{tabular}
  \end{center}
  \vspace{-3mm}
  \caption{\textbf{Performance comparison on the V-COCO}. The `\emph{B}', `\emph{P}' and `\emph{Q}' denote bounding-box, point and query, respectively.}
  \vspace{-3mm}
  \label{tb:vcoco}
\end{table}

\noindent\textbf{Implementation Details.} We implement three versions of GEN-VLKT: the small version GEN-VLKT$_s$, the middle version GEN-VLKT$_m$ and the large version GEN-VLKT$_l$. The backbone is ResNet-50 for GEN-VLKT$_s$ and ResNet-101 for GEN-VLKT$_m$ and GEN-VLKT$_l$. $N$ for each decoder of the two branches is $3$ for GEN-VLKT$_s$ and GEN-VLKT$_m$ and $6$ for GEN-VLKT$_l$. The number of HOI categories $N_a$ is $600$ for HICO-Det and $263$ for V-COCO. We set the number of queries $N_q$ to $64$ and the number of channels $C_e$ and $C_q$ to $256$. We optimize our network with AdamW with a weight decay of $10^{-4}$. We train the model for 90 epochs with an initial learning rate of $10^{-4}$ decreased by $10$ times at the $60$th epoch. The training is initialized with the parameters of MS-COCO trained DETR~\cite{carion2020endtoend}. For regular HOI training, we fine-tune the CLIP text embeddings initialized interaction classifier and object classifier with a small learning rate of $10^{-5}$. We implement the zero-shot HOI experiments on HICO-Det. For better novel HOI categories extension, we freeze the CLIP initialized weights for both interaction and object classifiers. We set the output dimension of the interaction classifier to the number of `seen' categories during training, while we update this output dimension to the `full' $600$ categories during inference. We set the cost weights $\lambda_b$, $\lambda_u$, $\lambda_c^o$ and $\lambda_c^a$ to $2.5$, $1$, $1$ and $1$, respectively, following QPIC~\cite{tamura2021qpic}. We follow the official CLIP data pre-processing for visual embedding mimic to resize and center-crop the real-timely augmented image to $224$ and feed the processed image to the CLIP visual encoder. We set the loss weight $\lambda_{mimic}$ to $20$. We conduct all the experiments with a batch size of $16$ on $8$ Tesla V100 GPUs and CUDA10.2.

\begin{table}[t]
\small
  \begin{center}
  \begin{tabular}{ccccc}
  \hline
  Method        &Type  & Unseen & Seen & Full \\
  \hline\hline
  Shen \emph{et al.}~\cite{shen2018scaling}      		&UC				    & 5.62  & - & 6.26\\
  FG~\cite{Bansal2020_aaai_functional}      		&UC				    & 10.93  & 12.60 & 12.26\\
  ConsNet~\cite{liu2020consnet}      		&UC				    & 16.99  & 20.51 & 19.81\\
  \hline
  VCL~\cite{hou2020visual}      		&RF-UC				    & 10.06  & 24.28 & 21.43\\
  ATL~\cite{hou2021affordance}      		&RF-UC				    & 9.18  & 24.67 & 21.57\\
  FCL~\cite{hou2021detecting}      		&RF-UC				    & 13.16  & 24.23 & 22.01\\
  \cellcolor{mygray-bg}baseline			 	&\cellcolor{mygray-bg}RF-UC			        &\cellcolor{mygray-bg}12.52  &\cellcolor{mygray-bg}32.70  &\cellcolor{mygray-bg}28.66\\
  \cellcolor{mygray-bg}GEN-VLKT$_s$			 	&\cellcolor{mygray-bg}RF-UC			        &\cellcolor{mygray-bg}\textbf{21.36}  &\cellcolor{mygray-bg}\textbf{32.91}  &\cellcolor{mygray-bg}\textbf{30.56}\\
  \hline 
  VCL~\cite{hou2020visual}      		&NF-UC				    & 16.22  & 18.52 & 18.06\\
  ATL~\cite{hou2021affordance}      		&NF-UC				    & 18.25  & 18.78 & 18.67\\
  FCL~\cite{hou2021detecting}      		&NF-UC				    & 18.66  & 19.55 & 19.37\\
  \cellcolor{mygray-bg}baseline		 	&\cellcolor{mygray-bg}NF-UC			        &\cellcolor{mygray-bg}18.71  &\cellcolor{mygray-bg}22.53  &\cellcolor{mygray-bg}21.76\\
  \cellcolor{mygray-bg}GEN-VLKT$_s$			 	&\cellcolor{mygray-bg}NF-UC			        &\cellcolor{mygray-bg}\textbf{25.05}  &\cellcolor{mygray-bg}\textbf{23.38}  &\cellcolor{mygray-bg}\textbf{23.71}\\
  \hline 
  FCL$^*$~\cite{hou2021detecting}      		&UO				    & 0.00  & 13.71 & 11.43\\
  ATL$^*$~\cite{hou2021affordance}      		&UO				    & 5.05  & 14.69 & 13.08\\
  \cellcolor{mygray-bg}baseline			 	&\cellcolor{mygray-bg}UO			        &\cellcolor{mygray-bg}2.92  &\cellcolor{mygray-bg}28.56  &\cellcolor{mygray-bg}23.99\\
  \cellcolor{mygray-bg}GEN-VLKT$_s$			 	&\cellcolor{mygray-bg}UO			        &\cellcolor{mygray-bg}\textbf{10.51}  &\cellcolor{mygray-bg}\textbf{28.92}  &\cellcolor{mygray-bg}\textbf{25.63}\\
  \hline
  \cellcolor{mygray-bg}baseline			 	&\cellcolor{mygray-bg}UV			        &\cellcolor{mygray-bg}13.52  &\cellcolor{mygray-bg}29.25  &\cellcolor{mygray-bg}27.04\\
  \cellcolor{mygray-bg}GEN-VLKT$_s$			 	&\cellcolor{mygray-bg}UV			        &\cellcolor{mygray-bg}\textbf{20.96}  &\cellcolor{mygray-bg}\textbf{30.23}  &\cellcolor{mygray-bg}\textbf{28.74}\\
  \hline 
  \end{tabular}
  \end{center}
  \vspace{-3mm}
  \caption{\textbf{Performance comparison for Zero-Shot HOI detection}. RF is short for rare first, NF is short for non-rare first, and UC, UO, UV indicate unseen composition, unseen object and unseen verb settings, respectively. The baseline is the model of `$s$' architecture without VLKT. $^*$ means only the detected boxes are used without object identity information from the detector.}
  \vspace{-3mm}
  \label{tb:zero-shot}
\end{table}

\vspace{-1.5mm}\subsection{Effectiveness for Regular HOI Detection}\label{sec:regular}\vspace{-1.5mm}

We use the official evaluation code to compute the mAPs for both HICO-Det and V-COCO. Table~\ref{tb:hico} and Table~\ref{tb:vcoco} show the performance comparisons of GEN-VLKT with the recent bottom-up and top-down HOI detection methods. 

\begin{table*}[t]
\begin{center}
\subfloat[\textbf{Network Architecture Setting}: Training `$s$' model without VLKT.\label{tab:ablation:arc}]{
\tablestyle{4pt}{1.05}\begin{tabular}{c|x{20}x{20}x{32}}
 Setting & Full & Rare & Non-Rare \\
\shline
\scriptsize Base-triplet & 30.96 & 22.28 & 33.55\\
  \scriptsize Base-verb& 31.88 &26.24 &33.57 \\
 \scriptsize \emph{- p-GE} &  31.23 & 25.38 & 32.98 \\
 \scriptsize \emph{- i-GE} &  30.83 & 23.86 & 32.91
\end{tabular}}\hspace{4mm}
\subfloat[\textbf{Training Strategies of VLKT.} : Ablations for the training strategies of VLKT. \label{tab:ablation:clip}]{
\tablestyle{4pt}{1.05}\begin{tabular}{c|x{20}x{20}x{32}}
  Strategy & Full & Rare & Non-Rare \\[.1em]
\shline
\scriptsize Base-triplet& 30.96 & 22.28 & 33.55\\
  \scriptsize \emph{+ interaction text}& 31.71 & 26.08 & 33.39 \\
  \scriptsize \emph{+ object text}&32.09&26.68&33.71 \\
  \scriptsize \emph{+ mimic}& \bd{33.75}&\bd{29.25}&\bd{35.10} 
\end{tabular}}\hspace{4mm}
\subfloat[\textbf{Mimic Loss Setting}: The choice of losses for mimic. \label{tab:ablation:mimic}]{
\tablestyle{4pt}{1.05}\begin{tabular}{cc|x{20}x{20}x{32}}
  $L_1$ & $L_2$ & Full &Rare&Non-Rare \\[.1em]
\shline
- & - & 32.09&26.68&33.71\\
\checkmark & - & \bd{33.75}&\bd{29.25}&\bd{35.10}\\
  - & \checkmark& 32.41 & 26.66& 34.14\\
  \checkmark & \checkmark& 33.10& 29.24 & 34.25
\end{tabular}}

\end{center}
\vspace{-4mm}
\caption{\textbf{Ablations}. We conduct experiments on HICO-Det dataset based on `$s$' model. The mAP in default setting is reported.}
\label{tab:ablations}\vspace{-3mm}
\end{table*}

For HICO-Det as shown in Table~\ref{tb:hico}, GEN-VLKT$_s$ outperforms the all existing bottom-up and top-down methods with a large margin. In specific, compared to the state-of-the-art top-down method QPIC~\cite{tamura2021qpic}, GEN-VLKT$_s$ achieves a relative  $16.10\%$ mAP gain with a margin of mAP $4.68$. Especially for the rare categories, GEN-VLKT$_s$ achieves mAP $29.25$, which significantly outperforms AS-Net~\cite{chen_2021_asnet} by a margin of mAP $5.00$, even outperforming the `Full' setting of all existing methods. This is ascribable to the guided embedding design of the GEN architecture and the powerful VLKT for the long-tail categories. From an efficiency perspective, GEN-VLKT$_s$ contains in total $6$ decoder layers considering the two branches. Thus it has almost the same number of parameters and flops compared to QPIC, and fewer parameters and flops compared to AS-Net~\cite{chen_2021_asnet} with $12$ decoder layers in total. And GEN-VLKT$_l$ achieves a new state-of-the-art performance of mAP $34.95$.

For V-COCO, as shown in Table~\ref{tb:vcoco}, GEN-VLKT$_s$ achieves role AP $62.41$ on Scenario $1$ and role AP $64.46$ on Scenario $2$, which also outperform the previous state-of-the-art method QPIC-R50~\cite{tamura2021qpic} with margins of mAP $3.61$ and $3.46$, respectively. 
The promotion is not as significant as that on HICO-Det, since the training samples of V-COCO is insufficient compared to HICO-Det to train such a large number of $263$ categories classification.

\vspace{-1.5mm}\subsection{Effectiveness for Zero-Shot HOI Detection}\label{sec:zero-shot}\vspace{-1.5mm}

We conduct all the experiments with the `$s$' model. We train the model without VLKT strategy as the baseline. As shown in Table~\ref{tb:zero-shot}, GEN-VLKT$_s$ outperforms the baseline and previous methods for all the Unseen Composition (UC), Unseen Object~(UO) and Unseen Verb~(UV) settings. 

\vspace{1mm}\noindent\textbf{Unseen Composition.} For UC, compared with FCL~\cite{hou2021detecting}, GEN-VLKT$_s$ achieves $38.85\%$ and $22.41\%$ relative mAP gains on full categories for rare first and non-rare first selections, respectively. Specifically, benefiting from the VLKT mechanism, GEN-VLKT$_s$ still significantly promotes the performance for the unseen categories without the feature factorization and composition among images like VCL~\cite{hou2020visual}, FCL~\cite{hou2021detecting} and ATL~\cite{hou2021affordance}. The improvements mainly come from the utilization of CLIP, as indicated by comparing GEN-VLKT$_s$ to the baseline. For example, for the rare first UC setting, GEN-VLKT$_s$ promotes mAP from $13.16$ to $21.36$ compared to FCL and promotes mAP by a significant margin of mAP $8.84$ compared to the baseline. 

\vspace{0.5mm}\noindent\textbf{Unseen Object.} We further evaluate GEN-VLKT$_s$ with unseen object, which reflects the ability to investigate human interactions with novel objects. For full and unseen categories, GEN-VLKT$_s$ outperforms ATL~\cite{hou2021affordance} by relative $95.95\%$ and $108.12\%$ mAP gains, respectively. Again, the comparison to the baseline indicates VLKT promotes the performance for the unseen categories significantly.

\vspace{0.5mm}\noindent\textbf{Unseen Verb.} Finally we propose the UV setting to discovery novel categories of actions, and we argue this reflects the specific characteristic of zero-shot HOI detection. We compare GEN-VLKT$_s$ with the baseline and obtain a significant $55.03\%$ relative promotion for unseen categories.

\vspace{-1.5mm}\subsection{Ablation Study}\label{sec:ablation}\vspace{-1.5mm}
In this subsection, we conduct a series of experiments to analyse the effectiveness of our proposed modules and strategies. All experiments are conducted in HICO-Det dataset based on the `$s$' model.

\vspace{0.5mm}\noindent\textbf{Network Architecture Setting.} In this part, we aim to prove the superiority of our designed framework. Thus, we implement two base models with regular training strategies without VLKT. Firstly, we follow the previous query-based methods to adopt a verb classifier with $117$ categories for GEN, namely `Base-verb' in Table~\ref{tab:ablation:arc}. It shows that our `Base-verb' has achieved $31.88$ mAP outperforming all existing HOI detectors. Secondly, we replace the verb classifier with an HOI triplet classifier with $600$ categories, namely `Base-triplet'. Due to the serious long-tailed distribution, it drops a bit mAP compared with `Base-Verb', especially for `Rare' HOIs.
Additionally, we explore the importance of components in our GEN. On the one hand, we remove p-GE and replace the independent human and object queries with a unified query for two tasks in `Base-verb', and the result has dropped a $0.65$ mAP. On the other hand, we remove i-GE and use a learnable embedding with random initialization added by the p-GE for interaction queries in `Base-verb'. It has lost $1.05$ mAP compared with `Base-verb', but still superior to previous single-branch methods. 

\vspace{0.5mm}\noindent\textbf{Training Strategies of VLKT.} Here, we verify the effectiveness of the proposed components in VLKT and the results are presented in Table~\ref{tab:ablation:clip}. We take the `Base-triplet' as the baseline model. We first replace the interaction classifier with the text embedding initialization, and the results are reported as `\emph{+interaction text}', which shows the mAP of `Rare' HOIs has improved a lot. Thus, the prior knowledge from linguistic prior can alleviate the long-tail distribution.
We then further equip this model with the text embedding driven object classifier, causing $0.38$ mAP improvement. Finally, we add the mimic loss to transfer visual knowledge from CLIP. All performances have boosted a lot, which proves aligning features is essential. 

\vspace{0.5mm}\noindent\textbf{Mimic Loss Setting.} We discuss the choice of the mimic loss with two loss types, \emph{i.e.}, $\mathcal{L}_1$ and $\mathcal{L}_2$ losses. As shown in Table~\ref{tab:ablation:mimic}, if only equipped with $\mathcal{L}_1$ loss, our model has achieved the best performance and a $1.66$ mAP gain, and the performance is much better than only equipped with $\mathcal{L}_2$ loss. If we apply $\mathcal{L}_1$ and $\mathcal{L}_2$ losses at the same time by summation, the performance is also worse than only with $\mathcal{L}_1$. Thus, $\mathcal{L}_1$ is more suitable for the mimic loss.  

\vspace{-2.5mm}\section{Conclusion}\vspace{-2mm}
We propose a novel framework GEN-VLKT to improve the query-based HOI detectors from two aspects, association and interaction understanding. For association, we design a two-branch framework while removing post-matching by a guided embedding mechanism. For interaction understanding, we design a training strategy VLKT, adopting CLIP to enhance interaction understanding. GEN-VLKT has achieved leading performances on regular and zero-shot settings on HICO-Det and V-COCO datasets. We conduct a simple attempt to CLIP, where only adopting a global knowledge distillation for visual features mimic may not be sufficient for a dense understanding task. In future, we aim further to mine the benefits of CLIP for HOI detection and improve the mimic strategy.

{\small
\bibliographystyle{ieee_fullname}
\bibliography{egbib}
}

\end{document}